%% file: root.tex
\documentclass[letterpaper, 10 pt, conference]{ieeeconf} 
\IEEEoverridecommandlockouts

\usepackage{cite}
\usepackage{amsmath,amssymb,amsfonts}
\usepackage{booktabs}
\usepackage{subcaption}
\usepackage{algorithmic}
\usepackage{graphicx}
\usepackage{textcomp}
\usepackage{comment}
\usepackage{xcolor}
\usepackage{multirow}

\begin{document}

\title{\LARGE \bf Dual Quaternion Based Contact Modeling for Fast and Smooth Collision Recovery of Quadrotors} 

\author{Valentin Gaucher and Wenlong Zhang%
\thanks{This work is supported by National Science Foundation under Grant No. 2331781.}%
\thanks{The authors are with School of Manufacturing Systems and Networks, Ira A. Fulton Schools of Engineering, Arizona State University, Mesa, AZ, 85212, USA. Email: \{{\tt vgaucher, wenlong.zhang\}@asu.edu}.}%
}

\maketitle
\thispagestyle{empty} 
\pagestyle{empty}    
\overrideIEEEmargins  

\begin{abstract}
Unmanned aerial vehicles (UAVs) operating in cluttered environments require \textcolor{black}{efficient and} accurate impact modeling to maintain stability post collisions, however \textcolor{black}{classical impulse} contact models decouple \textcolor{black}{the normal and tangential components}. This letter presents a dual quaternion \textcolor{black}{impulse} reset map directly on the $SE(3)$ manifold. By operating on the unified spatial twist (unified linear and angular velocities), the proposed formulation \textcolor{black}{retains the cross-coupling between normal and tangential impulse components in a single closed-form expression, and recovers the classical decoupled Newton impulse model as a special case}. A recovery controller is designed that couples linear and angular momentum to \textcolor{black}{enforce kinetic energy dissipation} across impacts. Hardware-in-the-loop benchmarks demonstrate a 24\% reduction in execution latency compared to an optimized matrix-based implementation\textcolor{black}{, and a 20\% reduction relative to a position-plus-quaternion (PQ) formulation}. MuJoCo simulations \textcolor{black}{across Monte Carlo sweeps over impact angles and friction coefficients show a 50.8\%--75.1\%} reduction in position root-mean-square error (RMSE) and a \textcolor{black}{68.7\%--85\%} decrease in peak kinetic energy \textcolor{black}{compared to published linear-admittance baselines}.
\end{abstract}

\vspace{1ex}

\input{01-Intro}
\input{02-sectionII}

\input{03-sectionIII}
\input{04-sectionIV}
\input{05-sectionV}

\bibliographystyle{IEEEtran}
\bibliography{IEEEabrv,bibliography}
\input{06-appendix}
\end{document}

%% file: 01-Intro.tex
\section{Introduction}
Unmanned aerial vehicles (UAVs) are increasingly deployed in cluttered environments that require collision recovery after physical interactions for stabilization \cite{Ruggiero} or tactile-based exploration \cite{patnaik2025}. In such settings, accurate modeling of impact-induced state discontinuities is essential for designing the recovery controllers. These collision events are inherently hybrid: the vehicle's states evolve continuously on the configuration manifold $SE(3)$ during free flight, and undergo instantaneous velocities jumps upon contact. Ensuring consistency of these discrete resets with the underlying geometric structure is critical for closed-loop stability.

\textcolor{black}{Matrix-based formulations represent the configuration manifold $SE(3)$ using rotation matrices $\mathbf{R} \in SO(3)$ and translation vectors $\mathbf{p} \in \mathbb{R}^3$. Classical rigid-body impact models, derived from Newton impulse-momentum principles, resolve a contact in these decoupled translational and rotational coordinates \cite{MarkMuller}. Such formulations determine the normal impulse magnitude from the contact normal alone and apply the tangential friction impulse separately, so the coupling between the normal and tangential components is not retained within the impulse magnitude. Resolving the two components jointly under Coulomb friction instead requires incremental or iterative numerical schemes \cite{CannyJohn,Stewart} that keep the impact dissipative \cite{Jia2011}. Position-plus-quaternion and exponential-coordinate representations \cite{BulloPDECC,GeometricSE3Control} reduce parameter redundancy but address the representation rather than the impact model, and still require explicit moment-arm computation or matrix exponentials within the control loop.}

To address this challenges, we develop a reset map directly on the dual quaternion manifold. The proposed formulation extends Newton’s restitution law to dual screws, \textcolor{black}{resolving the fully coupled $SE(3)$ cross-coupling in a single, closed-form operation. This embeds the physical coupling directly into the algebraic representation, providing an exact and computationally efficient alternative to traditional decoupled or iterative frameworks.}

\begin{figure}
    \centering
    \includegraphics[width=\linewidth]{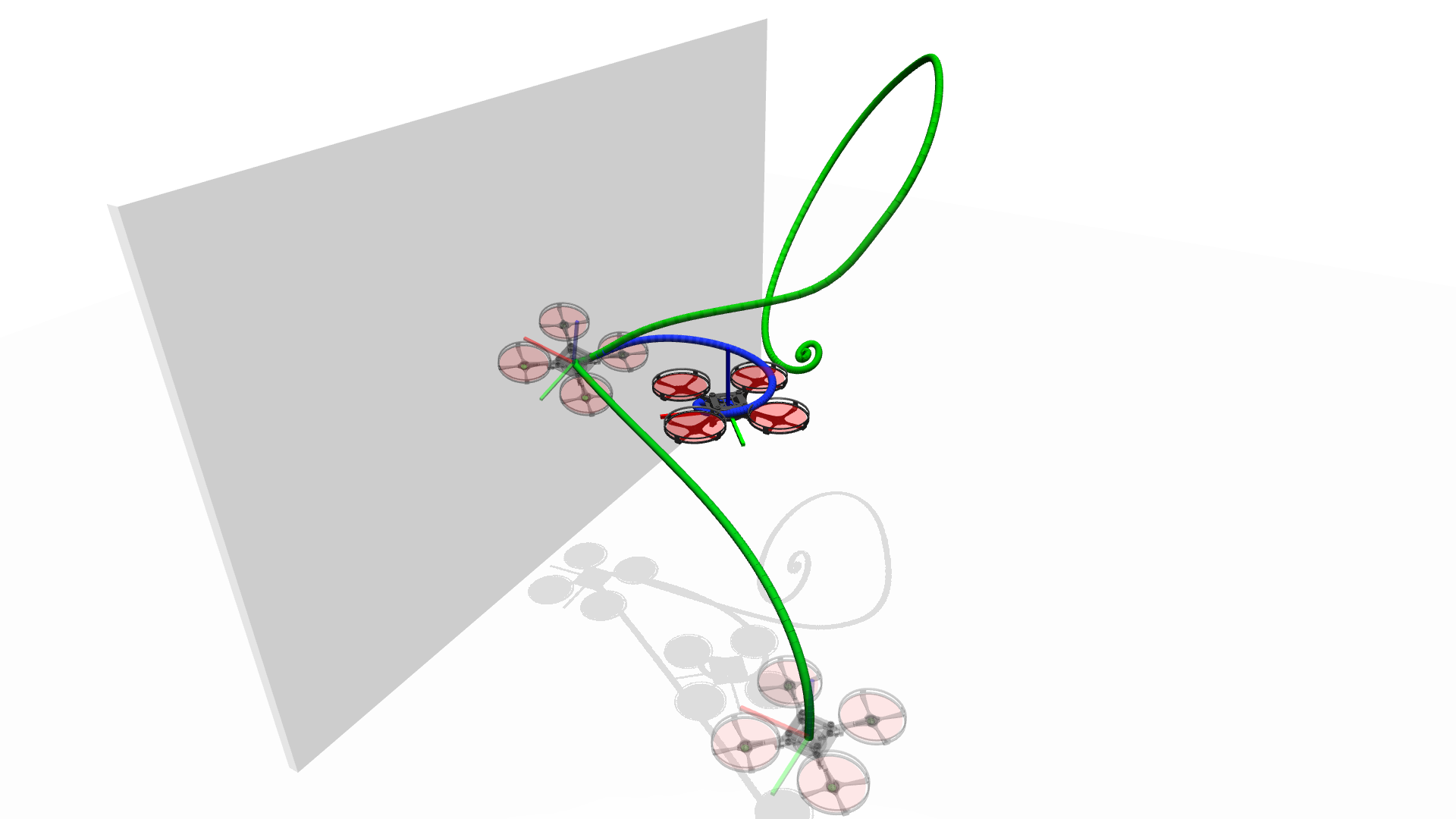}
    \caption{UAV collision recovery in MuJoCo simulation. The proposed dual quaternion framework (blue) mitigates linear and angular state oscillations post-impact, preventing drift seen in the traditional matrix-based approach (green).}
    \label{fig:frontfigure}
\end{figure}
\vspace{0.05in}
\noindent\textbf{Related Work.}
The mathematical foundation for rigid-body impacts is well-established, with the impulse-based collision resolution \cite{CannyJohn} providing the foundations for modern work such as a high-speed quadrotor recovery controller of \cite{MarkMuller}, which estimates post-impact states from a classical matrix-based formulation. Impacts are also modeled by solving linear complementarity problems (LCP) in the inertial frame \cite{Stewart} such as MuJoCo contact solving software \cite{todorov2012mujoco} and a recent LCP contact model for UAVs \cite{Abazari}. 

Dual quaternions are a powerful, unified algebraic alternative to matrix representations for rigid-body motion. Introduced by Clifford \cite{Clifford1871}, they map $SE(3)$ to the unit dual quaternion manifold and represent rotation and translation as a single screw displacement \cite{McCarthy}. Dual quaternion control laws avoid decoupled coordinate transformations, which makes them more computationally efficient and numerically stable than matrix-based formulations \cite{Wang2014,FilipeTsiotras}, and they have been applied to rigid-body trajectory tracking \cite{FilipeTsiotras,DualQuat2Quad,dqnmpc2025,Arrizabalaga}. Dual quaternions have not, however, been applied to the hybrid modeling of UAV collisions or to recovery control. Resolving an impact on this manifold allows the coupling between linear and angular momentum to be expressed within a single algebraic operation, rather than through separate translational and rotational updates as in the classical decoupled reset \cite{MarkMuller,Kazakov}.

\vspace{0.05in}
\noindent \textbf{Contributions.} \textcolor{black}{This letter makes three contributions. \textit{First}, we} derive a closed-form hybrid reset map on the dual quaternion representation of $SE(3)$ for rigid-body impacts. \textcolor{black}{The proposed expression (Eq.~(13)) retains the cross-coupling between normal and tangential impulse components in a single closed form and recovers the classical decoupled formulation as a special case (Proposition 1).} \textit{Second}, we develop a collision recovery controller \textcolor{black}{on the dual algebra} with proven hybrid Lyapunov stability \textcolor{black}{(Proposition 2), which integrates natively with $SE(3)$ feedback laws and standard cascaded UAV architectures}. \textit{Third}, we demonstrate that \textcolor{black}{the approach yields practical computational advantages, including a 25\% FLOP reduction over the matrix formulation and a 20\% reduction over the position-plus-quaternion formulation, validated by a 24\% latency reduction on embedded hardware. Closed-loop MuJoCo Monte Carlo simulations show improved recovery against published linear-admittance baselines, with the cross-coupling correction of Eq.~(13) producing observable transient improvement over the decoupled classical formulation.}

%% file: 02-sectionII.tex
\section{Preliminaries}

\subsection{Notation and Algebra}
Let $\mathcal{F}_W$ and $\mathcal{F}_B$ denote the inertial and body frames, respectively. The set of unit quaternions is $\mathbb{S}^3 \subset\mathbb{H}$ where $\mathbb{H}$ is the quaternion space. For any vector $\mathbf{v}\in\mathbb{R}^3$, we associate it with a pure quaternion in $\mathbb{H}_p$, and a rotation of $\mathbf{v}$ by $\mathbf{q}\in\mathbb{S}^3$ is written as $\mathbf{q}\odot \mathbf{v} = \mathbf{q} \otimes \mathbf{v} \otimes \mathbf{q}^*$, where $(\cdot)^*$ is the conjugate operation and $\otimes$ is the quaternion multiplication. The algebra of dual quaternions (DQ) $\mathcal{H}_d$ uses dual numbers $\hat{\mathbf{a}}=\mathbf{a}+\varepsilon\mathbf{b}$, where $\varepsilon$ is the nilpotent unit ($\varepsilon^2=0, \varepsilon \neq 0$). The dual operations dot product $\langle\cdot,\cdot\rangle$, cross product ($\times$ for kinematics) and adjoint cross product ($\times^*$ for kinetics) are:
\begin{align*}
    \langle \mathbf{a}+\varepsilon\mathbf{b}, \mathbf{c}+\varepsilon\mathbf{d} \rangle &= \mathbf{a}\cdot\mathbf{c} + \mathbf{b}\cdot\mathbf{d}\\
    (\mathbf{a}+\varepsilon\mathbf{b})\times (\mathbf{c}+\varepsilon\mathbf{d}) &= \mathbf{a}\times\mathbf{c} + \varepsilon(\mathbf{a}\times\mathbf{d}+\mathbf{b}\times\mathbf{c})\\
    (\mathbf{a}+\varepsilon\mathbf{b})\times^* (\mathbf{c}+\varepsilon\mathbf{d}) &= (\mathbf{a}\times\mathbf{c}+\mathbf{b}\times\mathbf{d}) + \varepsilon(\mathbf{a}\times\mathbf{d})
\end{align*}
For a dual matrix $\hat{\mathbf{K}} = \mathbf{A} + \varepsilon \mathbf{B}$, we use the decoupled matrix multiplication $\circ$ as $\hat{\mathbf{K}} \circ \hat{\mathbf{a}} =\mathbf{A}\mathbf{a} + \varepsilon \mathbf{B}\mathbf{b}$. These definitions can be found here \cite{FilipeTsiotras}, \textcolor{black}{and we adopt the dual kinematic convention of \cite{FilipeTsiotras}, where the primary and dual parts carry the rotational and translational components respectively and applied consistently, so that no swap operator is required.}

\subsection{Standard Quaternion Flight Dynamics}
The state of the UAV is defined by $(\mathbf{p}^W, \mathbf{v}^W, \mathbf{q}, \boldsymbol{\omega}^B)$, where $\mathbf{p}^W \in \mathbb{R}^3$ and $\mathbf{v}^W \in \mathbb{R}^3$ are the position and linear velocity in $\mathcal{F}_W$. $\mathbf{q} \in \mathbb{S}^3$ is the unit quaternion representing the orientation of $\mathcal{F}_B$ \textcolor{black}{and $\boldsymbol{\omega}^B \in \mathbb{R}^3$ is the angular velocity in $\mathcal{F}_B$ both relative to $\mathcal{F}_W$, represented in $\mathcal{F}_B$}. The continuous-time dynamics \textcolor{black}{use in the research literature \cite{ScaramuzzaEq}} are:
\begin{align}
\begin{split}
\dot{\mathbf{p}}^W = \mathbf{v}^W,& \quad \dot{\mathbf{q}} = \frac{1}{2} \mathbf{q} \otimes \boldsymbol{\omega}^B \\
\dot{\mathbf{v}}^W = g\mathbf{e}_z^W - \frac{f}{m} \mathbf{q}\odot\mathbf{e}_z^B,& \ \mathbf{J}\dot{\boldsymbol{\omega}}^B = \boldsymbol{\tau}^B - \boldsymbol{\omega}^B \times (\mathbf{J}\boldsymbol{\omega}^B)
\end{split}
\end{align}
where $m \in \mathbb{R}^{+}$ is the mass, \textcolor{black}{$g$ is the gravitational acceleration, $\mathbf{e}_z = [0, 0, 1]^\top$ is the vertical basis vector}, $\mathbf{J} \in \mathbb{R}^{3 \times 3}$ is the constant positive-definite inertia matrix, $f \in \mathbb{R}$ is the total thrust, and $\boldsymbol{\tau}^B \in \mathbb{R}^3$ is the control torque in $\mathcal{F}_B$.

\subsection{Matrix-Based Reset Model}
Here, we review the standard impulse-momentum impact model \cite{MarkMuller,CannyJohn}. Let the impact occur at point $\mathbf{r}^B_c$ (body frame) with normal $\mathbf{n}^W$ (world frame). \textcolor{black}{Given a pre-impact state ($\mathbf{v}^{W-},\ \boldsymbol{\omega}^{B-}$), the post-impact state ($\mathbf{v}^{W+},\ \boldsymbol{\omega}^{B+}$)} is given by the mapping $\mathcal{R}$:
\begin{equation}\label{eq:reset-SSM}
    \mathcal{R}:
  \begin{cases}
    \mathbf{v}^{W+} & = \mathbf{v}^{W-} + m^{-1} \boldsymbol{\lambda}^W \\
    \boldsymbol{\omega}^{B+} & = \boldsymbol{\omega}^{B-} + \mathbf{J}^{-1}(\mathbf{r}^B_c\times(\mathbf{R}^{\top} \boldsymbol{\lambda}^W)) 
  \end{cases}
\end{equation}
where $\boldsymbol{\lambda}^W$ is the impulse vector and $\mathbf{R}=\mathbf{R}(\mathbf{q})$ is the rotation matrix. \textcolor{black}{By expressing the contact normal in the body frame as $\mathbf{n}^B = \mathbf{R}^\top \mathbf{n}^W$}, the impulse magnitude is derived from the effective mass $\rho$ \textcolor{black}{(accounting for translational mass and the rotational moment-arm contribution)} as follows:
\begin{subequations}\label{eq:SMM}
\begin{align}\label{eq:standard_imp}
    &\boldsymbol{\lambda}^W = -\frac{(1+e) (\mathbf{v}_c^{W-})^\top \mathbf{n}^W}{\rho}(\mathbf{n}^W+\mu\, \mathbf{t}^W)\\
    \rho & = m^{-1} + \textcolor{black}{(\mathbf{r}^B_c \times \mathbf{n}^B)^\top \mathbf{J}^{-1} (\mathbf{r}^B_c \times \mathbf{n}^B)}
\end{align}
\end{subequations}
where $\mathbf{v}^{W-}_c = \mathbf{v}^{W-} + \mathbf{R}(\boldsymbol{\omega}^{B-}\times \mathbf{r}_c^B)$ is the pre-impact velocity, $e$ the restitution coefficient and $\mu$ the friction coefficient. The vectors $\mathbf{n}^W$ and $\mathbf{t}^W$ are the normal and tangential vectors at the impact point in $\mathcal{F}_W$.\\
\textcolor{black}{Note that Eq~(\ref{eq:SMM}) relies on the classical assumption that the normal and tangential impulse components decouple \cite{CannyJohn,Stewart,Jia2011}. Although this simplifies complex frictional dynamics, it provides a sufficient approximation for collision recovery, where the objective is rapid dissipation of kinetic energy rather than exact trajectory matching across the discrete impact event.}

%% file: 03-sectionIII.tex
\section{Dual Quaternion formulation}

\subsection{Dual Quaternion Flight Dynamics}
We represent the UAV configuration on $SE(3)$ using a unit dual quaternion $\hat{\mathbf{q}} \in \mathcal{H}_d$, providing a global, non-singular mapping of the translational position $\mathbf{p} \in \mathbb{R}^3$ and orientation $\mathbf{q} \in \mathbb{S}^3$. The velocities are encapsulated by the dual twist $\hat{\boldsymbol{\xi}} \in \mathcal{H}_d^p$, a pure dual quaternion composed of the body-frame angular velocity $\boldsymbol{\omega}^B \in \mathbb{R}^3$ and linear velocity $\mathbf{v}^B \in \mathbb{R}^3$. We formalize 
the dual dynamics, denoted $F$, below \cite{DualQuat2Quad,dqnmpc2025}
\begin{align}\label{eq:F}
\begin{split}
    \mathbf{\hat{q}}  &= \mathbf{q} + \varepsilon (\frac{1}{2}\mathbf{p}\otimes \mathbf{q}),\quad
    \boldsymbol{\hat{\xi}}  = \boldsymbol{\omega}^B + \varepsilon \mathbf{v}^B,\\
    \dot{\hat{\mathbf{q}}}  &= \frac{1}{2}\mathbf{\hat{q}} \otimes \boldsymbol{\hat{\xi}},\quad
    \boldsymbol{\dot{\hat{\xi}}}  = \mathcal{M}^{-1}(\hat{\mathbf{F}} - \boldsymbol{\hat{\xi}} \times^* \hat{\mathbf{H}} ), 
\end{split}
\end{align}
where \textcolor{black}{$\mathbf{v}^B = \mathbf{q}^*\odot\mathbf{v}^W$} and $\hat{\mathbf{H}}$ is the dual momentum \cite{Brodsky} defined via the dual inertia operator $\mathcal{M}: \mathcal{T}_{\hat{\mathbf{q}}}\mathcal{H}_d \rightarrow  \mathcal{T}^*_{\hat{\mathbf{q}}}\mathcal{H}_d$ mapping the tangent space (velocities) to the cotangent space (forces/momenta) at the center of mass.
Applying this operator to the twist gives the dual momentum:
\begin{equation}
    \hat{\mathbf{H}} = \mathcal{M}(\hat{\boldsymbol{\xi}}) = \mathbf{J}\boldsymbol{\omega}^B + \varepsilon (m \mathbf{v}^B)
\end{equation}
In the body frame, the operator $\mathcal{M}$ is constant and its inverse is exactly defined as:
\begin{equation}
    \mathcal{M}^{-1}(\hat{\boldsymbol{\xi}}) = \mathbf{J}^{-1}\boldsymbol{\omega}^B + \varepsilon(m^{-1}\mathbf{v}^B).
\end{equation}
$\hat{\mathbf{F}}=\hat{\mathbf{F}}_{a} + \hat{\mathbf{F}}_{g}$ represents the total external dual wrench acting on the UAV CoM, expressed in the body frame. $\hat{\mathbf{F}}_{a}$ (thrust and torque) and the gravitational wrench $\hat{\mathbf{F}}_{g}$ are written as
\begin{equation}\label{eq:command}
    \hat{\mathbf{F}}_{g} = \mathbf{0} + \varepsilon\mathbf{q}^*\odot(m\mathbf{g}^W),\quad
     \hat{\mathbf{F}}_{a} = \boldsymbol{\tau} + \varepsilon\boldsymbol{f},
\end{equation}
with \textcolor{black}{$\mathbf{0}=[0,0,0]\in\mathbb{R}^3$}, $\boldsymbol{f}\in\mathbb{R}^3$ the thrust vector and $\mathbf{g}^W=g\, \mathbf{e}^W_z\textcolor{black}{=[0,0,g]\in\mathbb{R}^3}$.

\subsection{The Dual Quaternion Reset Map}
\textbf{Assumption 1.} The collision between the UAV and the environment is modeled as a perfectly rigid impact over an infinitesimally small time interval. The configuration of the UAV remains continuous across the impact event ($\hat{\mathbf{q}}^+ = \hat{\mathbf{q}}^-$).

\begin{figure}[h]
    \centering
    \includegraphics[width=1.0\linewidth]{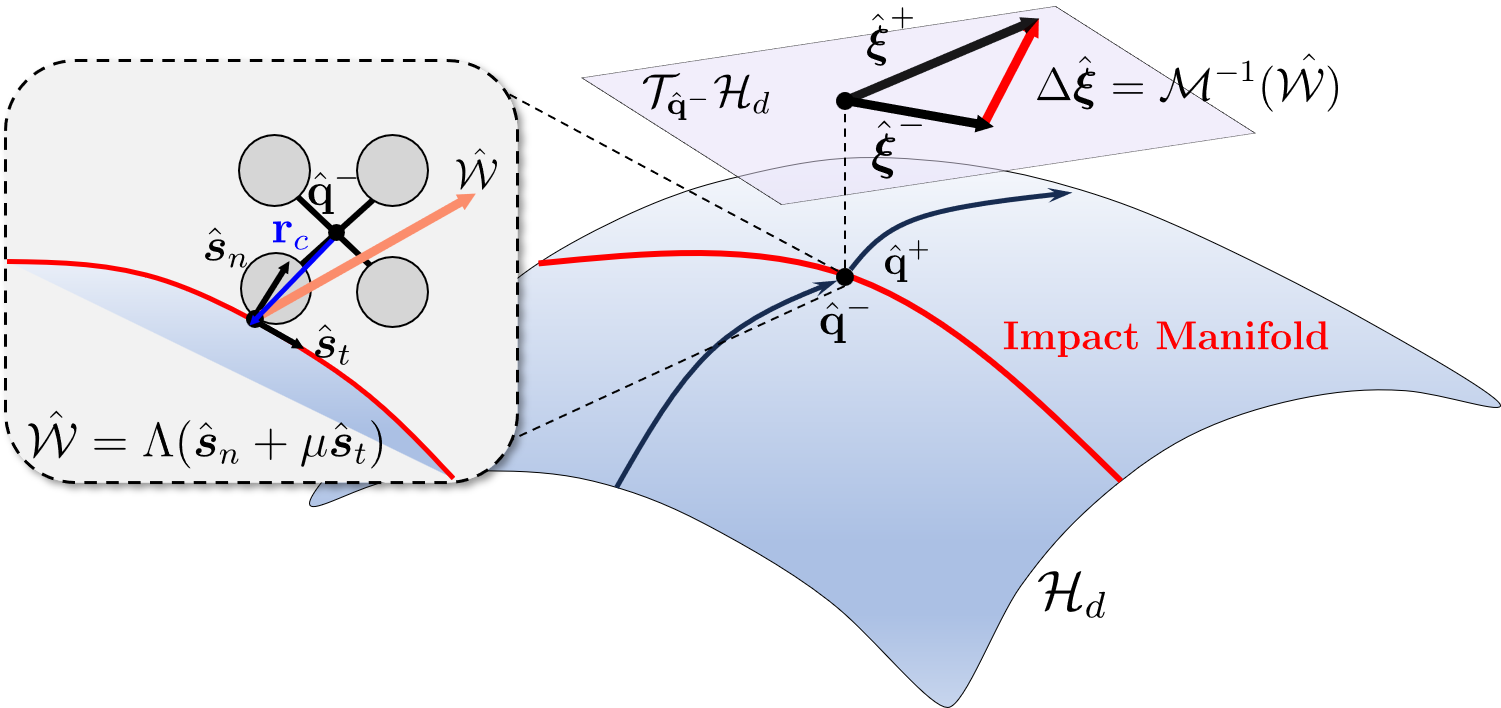}
    \caption{Dual quaternion reset map.}
    \label{fig:reset}
\end{figure}

We propose a DQ reset map $\hat{\mathcal{R}}$ that operates on the twist $\hat{\boldsymbol{\xi}}$. Unlike classical formulations that require intermediate mappings out of the configuration space to $SO(3)$ rotation matrices to resolve collisions, our approach evaluates the impact within the dual algebra. This preserves the $SE(3)$ geometry across the discrete jump while satisfying the physics. Under Assumption 1, the impact law relates the discrete jump in dual momentum $\hat{\mathbf{H}}$ to the external impulsive dual wrench $\hat{\mathcal{W}}$ acting on the system as follows:
\begin{equation}
    \Delta \hat{\mathbf{H}} = \hat{\mathbf{H}}^+ - \hat{\mathbf{H}}^- = \hat{\mathcal{W}}.
\end{equation}
Substituting the algebraic relationship $\hat{\mathbf{H}} = \mathcal{M}(\hat{\boldsymbol{\xi}})$, we obtain the following DQ reset map (see Fig.~\ref{fig:reset}):
\begin{equation}\label{eq:DQM-reset}
    \hat{\mathcal{R}}: \boldsymbol{\hat{\xi}}^+ = \boldsymbol{\hat{\xi}}^- + \mathcal{M}^{-1}(\hat{\mathcal{W}}).
\end{equation}

\noindent The impact geometry is defined by the contact point $\mathbf{r}_c^B$ and the surface normal $\mathbf{n}^W\in\mathbb{R}^3$. We transform the world-frame normal to the body frame using the current attitude $\mathbf{q}$ with $\mathbf{n}^B = \mathbf{q}^*\odot \mathbf{n}^W$. We define the collision geometry using two orthogonal unit screws in Plücker line coordinates \cite{McCarthy}, the normal screw $\hat{\mathbf{s}}_n\in\mathcal{H}_d$ (penetration) and the tangential screw $\hat{\mathbf{s}}_t\in\mathcal{H}_d$ (slip):
\begin{equation}\label{eq:Plucker}
    \hat{\mathbf{s}}_n =  (\mathbf{r}_c^B \times \mathbf{n}^B)+\varepsilon\mathbf{n}^B,\quad
    \hat{\mathbf{s}}_t = (\mathbf{r}_c^B \times \mathbf{t}^B) + \varepsilon\mathbf{t}^B,
\end{equation}
where $\mathbf{t}^B$ is the unit vector in the direction of the sliding velocity. We model the total collision wrench $\hat{\mathcal{W}}$ by applying a normal impulse (magnitude $\Lambda$) and a tangential friction impulse ($F_t$) along their respective lines of action. Because the screws in (\ref{eq:Plucker}) are structured as Plücker lines, scaling them by scalar impulses generates both the 3D contact forces and their induced 3D torques about the center of mass. Using the Coulomb friction model with  $\mu$ (where $F_t = \mu \Lambda$ \textcolor{black}{is the sliding condition}), the dual wrench is
\begin{equation}\label{eq:wrench}
    \hat{\mathcal{W}}=\Lambda \hat{\mathbf{s}}_n + \mu \Lambda \hat{\mathbf{s}}_t =\Lambda(\hat{\mathbf{s}}_n + \mu \hat{\mathbf{s}}_t).
\end{equation}
A visual representation of the dual wrench is shown in Fig. \ref{fig:reset}. To determine the scalar magnitude $\Lambda$, we apply Newton’s Law of Restitution, \textit{i.e.}, the post-impact relative velocity along the normal direction $v_n^+$ is related to the pre-impact velocity $v_n^-$ by the coefficient of restitution $e$ with $v_n^+ = -e v_n^-$ \textcolor{black}{\cite{Kazakov}}. In dual algebra, the projection of a twist $\hat{\boldsymbol{\xi}}$ onto a screw axis $\hat{\mathbf{s}}$ is given by the scalar part of the dual dot product $ v_n = \langle \hat{\boldsymbol{\xi}}, \hat{\mathbf{s}}_n \rangle$. Projecting the DQ reset law \eqref{eq:DQM-reset} onto the normal screw $\hat{\mathbf{s}}_n$ yields
\begin{subequations}
\begin{align}
    \langle \hat{\boldsymbol{\xi}}^+, \hat{\mathbf{s}}_n \rangle &= \langle \hat{\boldsymbol{\xi}}^-, \hat{\mathbf{s}}_n \rangle + \langle \mathcal{M}^{-1}(\hat{\mathcal{W}}), \hat{\mathbf{s}}_n \rangle \\
  -e \langle \hat{\boldsymbol{\xi}}^-, \hat{\mathbf{s}}_n \rangle &= \langle \hat{\boldsymbol{\xi}}^-, \hat{\mathbf{s}}_n \rangle + \Lambda \langle \mathcal{M}^{-1}(\hat{\mathbf{s}}_n + \mu \hat{\mathbf{s}}_t), \hat{\mathbf{s}}_n \rangle
\end{align}
\end{subequations}
and we get the final closed-form \textcolor{black}{fully coupled} expression for the impulse magnitude $\Lambda$:
\textcolor{black}{
\begin{equation}\label{eq:DQM}
    \Lambda = \frac{-(1+e) \langle \hat{\boldsymbol{\xi}}^-, \hat{\mathbf{s}}_n \rangle}{\mu \langle \mathcal{M}^{-1}(\hat{\mathbf{s}}_t), \hat{\mathbf{s}}_n \rangle + \langle \mathcal{M}^{-1}(\hat{\mathbf{s}}_n), \hat{\mathbf{s}}_n \rangle}
\end{equation}
}
To determine the final post-impact state, the computed impulse magnitude $\Lambda$ from Eq. \eqref{eq:DQM} is substituted into Eq. \eqref{eq:wrench} to formulate the total collision wrench $\hat{\mathcal{W}}$. Applying this wrench to the update law in Eq. \eqref{eq:DQM-reset} yields the post-impact twist $\hat{\boldsymbol{\xi}}^+$, \textcolor{black}{and is further illustrated Fig. \ref{fig:reset}}. \textcolor{black}{In practice, collision parameters can be estimated with tactile sensing \cite{patnaik2025,MarkMuller,Tomic}. Pre-impact velocities are taken from onboard state estimation, and $(e,\mu)$ are calibrated offline \cite{MarkMuller,Chui2016}.}

\noindent\textbf{Proposition 1.} \textcolor{black}{Under the same decoupled assumption \cite{CannyJohn,Stewart,Jia2011} adopted in Eq.~(\ref{eq:SMM}),} the dual quaternion reset map is equivalent to the standard matrix reset map ($\mathcal{R}\equiv \hat{\mathcal{R}}$).

\vspace{0.05in}
\noindent\textbf{Proof.} First, consider the numerator of \eqref{eq:DQM}. The dual dot product $\langle \hat{\boldsymbol{\xi}}^-, \hat{\mathbf{s}}_n \rangle$ projects the twist onto the normal screw axis. This operation directly evaluates the linear velocity of the contact point along the collision normal, which is kinematically equivalent to the classical formulation:
\begin{equation}
\langle \hat{\boldsymbol{\xi}}^-, \hat{\mathbf{s}}_n \rangle \equiv (\mathbf{v}_c^{W-})^\top \mathbf{n}^W.
\end{equation}
Next, consider the denominator of Eq. \eqref{eq:DQM} denoted $\rho_{DQ}$. \textcolor{black}{To match the decoupled assumption \cite{CannyJohn,Stewart,Jia2011} adopted in Eq.~(\ref{eq:SMM}), we neglect the cross-coupling term:
\begin{equation}\label{eq:assumption}
    \mu \langle \mathcal{M}^{-1}(\hat{\mathbf{s}}_t), \hat{\mathbf{s}}_n \rangle \approx 0
\end{equation}   
}
Recalling the definition of the inverse dual inertia operator, we evaluate its action on the normal screw as $\mathcal{M}^{-1}(\hat{\mathbf{s}}_n) = \mathbf{J}^{-1}(\mathbf{r}_c^B \times \mathbf{n}^B) + \varepsilon (m^{-1} \mathbf{n}^B)$. Expanding the dual dot product $\rho_{DQ}$, with $\mathbf{n}{}^{B,\top} \mathbf{n}^B = 1$, yields:
\begin{equation}
    \rho_{DQ} = (\mathbf{r}_c^B \times \mathbf{n}^B)^{\top} \mathbf{J}^{-1}(\mathbf{r}_c^B \times \mathbf{n}^B) +  m^{-1}.
\end{equation}
\textcolor{black}{Given the effective inverse mass in Eq. \eqref{eq:SMM}, we have $\rho_{DQ} \equiv \rho$.} The effective collision inertia is identical. The discrete dual state update is $\Delta \hat{\boldsymbol{\xi}} \textcolor{black}{= \hat{\boldsymbol{\xi}}^+ - \hat{\boldsymbol{\xi}}^-}=\mathcal{M}^{-1}(\hat{\mathcal{W}})$. Using the total impulse in the body frame $\boldsymbol{\lambda}^B = \Lambda (\mathbf{n}^B + \mu \mathbf{t}^B)$, \textcolor{black}{along with Eq~(\ref{eq:Plucker})-(\ref{eq:wrench})}, the dual wrench $\hat{\mathcal{W}}$ can be developed as:
\begin{equation}
    \hat{\mathcal{W}} = (\mathbf{r}_c^B \times \boldsymbol{\lambda}^B) + \varepsilon \boldsymbol{\lambda}^B.
\end{equation}
Applying $\mathcal{M}^{-1}(\cdot)$ to $\hat{\mathcal{W}}$ gives the following twist update
\begin{equation}
    \Delta \hat{\boldsymbol{\xi}} = \mathbf{J}^{-1}(\mathbf{r}^B_c \times \boldsymbol{\lambda}^B) + \varepsilon \left( m^{-1} \boldsymbol{\lambda}^B \right) = \Delta \boldsymbol{\omega}^B + \varepsilon \Delta \mathbf{v}^B.
\end{equation}
The angular update $\Delta \boldsymbol{\omega}^{B}\textcolor{black}{=\boldsymbol{\omega}^{B+}-\boldsymbol{\omega}^{B-}}$ matches the standard model in \eqref{eq:reset-SSM} exactly. For the linear update $\Delta \mathbf{v}^{B}\textcolor{black}{=\mathbf{v}^{B+}-\mathbf{v}^{B-}}$, the standard model updates inertial velocity $\mathbf{v}^W$. The DQ model updates body velocity $\mathbf{v}^B$ as
\begin{equation}
    \Delta \mathbf{v}^W = \mathbf{R} (\Delta \mathbf{v}^B) = \mathbf{R} \left( m^{-1} \boldsymbol{\lambda}^B \right) = m^{-1} \boldsymbol{\lambda}^W.
\end{equation}
Since the reset update is expressed directly in the twist associated with the DQ representation, the post-impact state remains on the manifold without intermediate coordinate transformations. Therefore, the proposed reset map is mathematically equivalent to the classical model in \eqref{eq:reset-SSM}. $\blacksquare$

\vspace{0.05in}
\noindent \textcolor{black}{\textbf{Remark 1.} Classical reset implementations represent the post-impact velocity as a pair $(\Delta \mathbf{v}, \Delta \boldsymbol{\omega}) \in \mathbb{R}^3 \times \mathbb{R}^3$ (Eq. (\ref{eq:reset-SSM})), with the moment-arm contribution to the angular update computed through an explicit cross product $\mathbf{r}_c \times \boldsymbol{\lambda}$. The proposed reset map updates the dual twist $\hat{\boldsymbol{\xi}} \in SE(3)$ directly in dual algebra, with the cross-product term absorbed into the algebraic structure through the screw $\hat{\mathbf{s}}_n$. The post-impact state is therefore in the native format for $SE(3)$ feedback laws, without intermediate frame conversions.}

\subsection{Dual Quaternion Impact Recovery Control}
\label{sec:recovery}
To avoid actuator saturation during collision events, we implement an admittance-based recovery strategy directly on the DQ manifold, adapting the classical non-dual formulations seen in \cite{MarkMuller,Patnaik2021}. At impact time $t_c$, the closed-form impulse magnitude $\Lambda$ in (\ref{eq:DQM}) and impulsive wrench $\hat{\mathcal{W}}$ in (\ref{eq:wrench}) are used to compute the post-impact twist via the dual reset map in (\ref{eq:DQM-reset}). We define the following dual admittance gain $\hat{\mathbf{\Gamma}} \in \mathbb{R}^{3 \times 3} + \varepsilon \mathbb{R}^{3 \times 3}$
\begin{equation}
\hat{\mathbf{\Gamma}} = \mathbf{\Gamma}_\omega + \varepsilon \mathbf{\Gamma}_v,\quad \mathbf{\Gamma}_i = \text{diag}(\gamma_{i,x},\gamma_{i,y},\gamma_{i,y} ).
\end{equation}
Here, $\mathbf{\Gamma}_\omega, \mathbf{\Gamma}_v \in \mathbb{R}^{3 \times 3}$ are positive-definite diagonal matrices containing the independent control gains for the rotational and translational axes. We scale the post-impact twist to obtain the braking displacement in the Lie algebra as follows
\begin{equation}
\boldsymbol{\hat{\delta}} = \hat{\mathbf{\Gamma}} \circ \boldsymbol{\hat{\xi}}^+
= \mathbf{\Gamma}_\omega \boldsymbol{\omega}^+ + \varepsilon \mathbf{\Gamma}_v \mathbf{v}^{B+} .
\label{eq:dual-recovery}
\end{equation}
The displacement $\boldsymbol{\hat{\delta}}$ is mapped to the configuration manifold via the dual exponential $\hat{\mathbf{q}}_{\Delta} = \exp(\tfrac{1}{2}\boldsymbol{\hat{\delta}})$ \cite{dqnmpc2025}. The recovery setpoint $\hat{\mathbf{q}}_d$ and configuration error $\hat{\mathbf{q}}_e$ are defined as
\begin{subequations}\label{eq:dual_exp}
\begin{align}
\hat{\mathbf{q}}_d &= \hat{\mathbf{q}}_{\Delta} \otimes \hat{\mathbf{q}}(t_c)\, ,\\
\hat{\mathbf{q}}_e &= \hat{\mathbf{q}}_d^* \otimes \hat{\mathbf{q}} = \mathbf{q}_e + \varepsilon\frac{1}{2}\mathbf{p}_e \otimes \mathbf{q}_e\, . \label{eq:qehat}
\end{align}
\end{subequations}
\textcolor{black}{With $\hat{\mathbf{q}}_d$ fixed at $t_c$, the reference twist vanishes and the body twist $\hat{\boldsymbol{\xi}}$ is itself the velocity-error coordinate.} Shifting the reference \textcolor{black}{setpoint} introduces virtual compliance, allowing the UAV to yield to the collision.

\vspace{0.05in}
\noindent\textcolor{black}{\textbf{Remark 2.}
To prevent unwinding due to the $\mathbb{S}^3$ over $SO(3)$ structure, we enforce shortest-path rotation \cite{FilipeTsiotras} by mapping $\hat{\mathbf{q}}_e \leftarrow -\hat{\mathbf{q}}_e$ when the scalar real part $q_e^w < 0$ prior to control.}

\vspace{0.05in}
\noindent \textcolor{black}{ \textbf{Remark 3.} Our formulation extends the classical translational admittance of \cite{MarkMuller,Patnaik2021} to a unified $SE(3)$ screw shift. The dual exponential in Eq.~\eqref{eq:dual_exp} composes the linear and angular setpoint updates into a single screw displacement on $SE(3)$, in place of the two independent shifts on $\mathbb{R}^3 \times SO(3)$ required by a non-dual implementation.}

\vspace{0.05in}
\noindent\textbf{Hybrid Lyapunov Stability}
Defining the state $x = (\hat{\mathbf{q}}, \hat{\boldsymbol{\xi}})$, we model the closed-loop recovery system as a hybrid dynamical system $\mathcal{H} = (F, \mathcal{C}, \mathcal{D}, \hat{\mathcal{R}})$ satisfying standard basic conditions \cite{goebel2012hybrid}. The flow and jump sets are $\mathcal{C} = \{ x \mid \phi(\hat{\mathbf{q}}) > 0 \}$ and $\mathcal{D} = \{ x \mid \phi(\hat{\mathbf{q}}) \le 0 \}$, where $\phi$ is the signed distance to the contact surface. Since \textcolor{black}{$e \in [0,1)$} ensures bounded kinetic energy dissipation at each impact, Zeno executions are excluded. Using (\ref{eq:qehat}), with $q_e^w$ and $\mathbf{q}_e^v$ being the scalar and vector parts of $\mathbf{q}_e$, the continuous dynamics $F$ from Eq. (\ref{eq:F}) are governed by the control law:
\begin{equation}
\hat{\mathbf{F}}_a 
= -\hat{\mathbf{F}}_g 
+ \boldsymbol{\hat{\xi}} \times^* \hat{\mathbf{H}} 
- \underbrace{(k_q \mathbf{q}_{e}^v + \varepsilon k_p \mathbf{p}_e)}_{\hat{\mathbf{e}}} 
- \mathbf{K}_d \circ \boldsymbol{\hat{\xi}}
\label{eq:control_law}
\end{equation}
\textcolor{black}{with $k_p, k_q \in \mathbb{R}^+$ proportional and derivative control gains.} This yields nonlinear decoupling and dual damping ($\mathbf{K}_d > 0$). \textcolor{black}{Eq.~(\ref{eq:control_law}) is the wrench-level control law used for the stability analysis. In implementation its control intent is realized through the shifted reference of Remark~4.}

\vspace{0.05in}
\noindent \textcolor{black}{\textbf{Remark 4.} The recovery admittance map outputs only a shifted reference on $SE(3)$. Reference tracking is performed by a geometric controller \cite{GeometricSE3Control}. The desired force is realized as a thrust along body-z together with a desired attitude, and the resulting force–moment pair is mapped to four rotor angular velocities through the standard allocation matrix \cite{ScaramuzzaEq}. This inner loop is shared by all the compared methods.}

\vspace{0.05in}
\noindent\textbf{Proposition 2.}
The closed-loop hybrid system achieves Hybrid Lyapunov stability at $\mathcal{A} = \{ x \mid \hat{\mathbf{q}}_e = \hat{\mathbf{1}}, \boldsymbol{\hat{\xi}} = \hat{\mathbf{0}} \}$, \textcolor{black}{where $\hat{\mathbf{1}} = (1, \mathbf{0}) + \varepsilon (0, \mathbf{0})$, and $\hat{\mathbf{0}} = (0, \mathbf{0}) + \varepsilon (0, \mathbf{0})$.}

\vspace{0.05in}
\noindent\textbf{Proof.}
(Extended version is available in the Appendix).
Consider the candidate Lyapunov function,
\textcolor{black}{with $\| \boldsymbol{\hat{\xi}} \|_{\mathcal{M}}^2 =
\langle \boldsymbol{\hat{\xi}}, \mathcal{M}(\boldsymbol{\hat{\xi}})
\rangle$}:
\begin{equation}
V(x)
= \underbrace{2k_q(1 - q_{e}^w)
+ \frac{1}{2}k_p \|\mathbf{p}_e\|^2}_{V_{pos}}
+ \underbrace{\frac{1}{2}\textcolor{black}{\| \boldsymbol{\hat{\xi}}
\|_{\mathcal{M}}^2}}_{V_{kin}}
\end{equation}
positive definite on $\mathcal{A}$ \textcolor{black}{and well defined provided Remark~2}.\\
\noindent\textit{Flow condition ($x \in \mathcal{C}$).}
Since the reference pose is constant during free flight,
differentiating $V$ along the flow gives
\begin{equation}
\dot{V} = \langle \hat{\mathbf{e}}, \hat{\boldsymbol{\xi}} \rangle
+ \langle \hat{\boldsymbol{\xi}}, \hat{\mathbf{F}}_a + \hat{\mathbf{F}}_g
- \hat{\boldsymbol{\xi}} \times^* \hat{\mathbf{H}} \rangle.
\end{equation}
Substituting \eqref{eq:control_law} yields cancellation of
gravitational, gyroscopic, and potential gradient terms.
The remaining term is
\begin{equation}
\dot{V}
= - \langle \boldsymbol{\hat{\xi}}, \mathbf{K}_d \circ
\boldsymbol{\hat{\xi}} \rangle
\le 0
\end{equation}
establishing non-increasing energy along flows.\\
\noindent\textit{Jump condition ($x \in \mathcal{D}$).}
\textcolor{black}{At impact, Assumption~1 gives
$\hat{\mathbf{q}}^{+} = \hat{\mathbf{q}}^{-}$, so the actual
configuration does not change. The reference update
Eq.~(\ref{eq:dual_exp}) shifts the setpoint by
$\hat{\mathbf{q}}_\Delta$, yielding
$\Delta V_{pos} \leq V_{pos}(\hat{\mathbf{q}}_\Delta)$.}
The change in kinetic energy satisfies
\begin{equation}
\Delta V_{kin}
= \langle \boldsymbol{\hat{\xi}}^-, \hat{\mathcal{W}} \rangle
+ \frac{1}{2}
\langle \hat{\mathcal{W}}, \mathcal{M}^{-1}(\hat{\mathcal{W}}) \rangle.
\end{equation}
\textcolor{black}{Using} $\hat{\mathcal{W}}_n = \Lambda \hat{\mathbf{s}}_n$
\textcolor{black}{(Eq.~(\ref{eq:wrench}))} and the restitution relation:
\begin{equation}\label{eq:deltaV}
\Delta V_{kin,n}
= -\frac{1}{2}\Lambda^2
\langle \hat{\mathbf{s}}_n, \mathcal{M}^{-1}(\hat{\mathbf{s}}_n) \rangle
\left(\frac{1-e}{1+e}\right).
\end{equation}
Since $\mathcal{M} \succ 0$ and \textcolor{black}{$e \in [0,1)$},
Eq.~(\ref{eq:deltaV}) is strictly negative.
Coulomb friction ensures $\Delta V_{kin,t} \le 0$, yielding total
dissipation $\Delta V_{kin} = -E_{diss} < 0$.
The setpoint update introduces a bounded pose shift
$\hat{\mathbf{q}}_\Delta$. For braking gains
$(\boldsymbol{\Gamma}_\omega, \boldsymbol{\Gamma}_v)$ satisfying
\begin{equation}
\frac{1}{4} k_q \|\boldsymbol{\Gamma}_\omega \boldsymbol{\omega}^+\|^2
+
\frac{1}{2} k_p \|\boldsymbol{\Gamma}_v\mathbf{v}^{B+}\|^2 <
E_{diss} + V_{pos}(\hat{\mathbf{q}}_e^-),
\end{equation}
\textcolor{black}{the injected potential energy satisfies
$V_{pos}(\hat{\mathbf{q}}_\Delta) < E_{diss}$, so that
$\Delta V \leq V_{pos}(\hat{\mathbf{q}}_\Delta) - E_{diss} < 0$.}
Since $\dot{V} \le 0$ on $\mathcal{C}$ and $\Delta V < 0$ on
$\mathcal{D}$, Lyapunov stability of the closed-loop hybrid system
follows \cite{goebel2012hybrid}. \hfill $\blacksquare$. \textcolor{black}{For selecting $(\boldsymbol{\Gamma}_v,
\boldsymbol{\Gamma}_\omega)$, we refer the reader to the extended
version of the proof
in the Appendix.}

%% file: 04-sectionIV.tex
\section{Numerical Validation and Simulation}
\label{sec:numerical_results}
\begin{figure}[h]
    \centering
    \begin{subfigure}{1.0\linewidth}
        \centering
        \includegraphics[width=\linewidth]{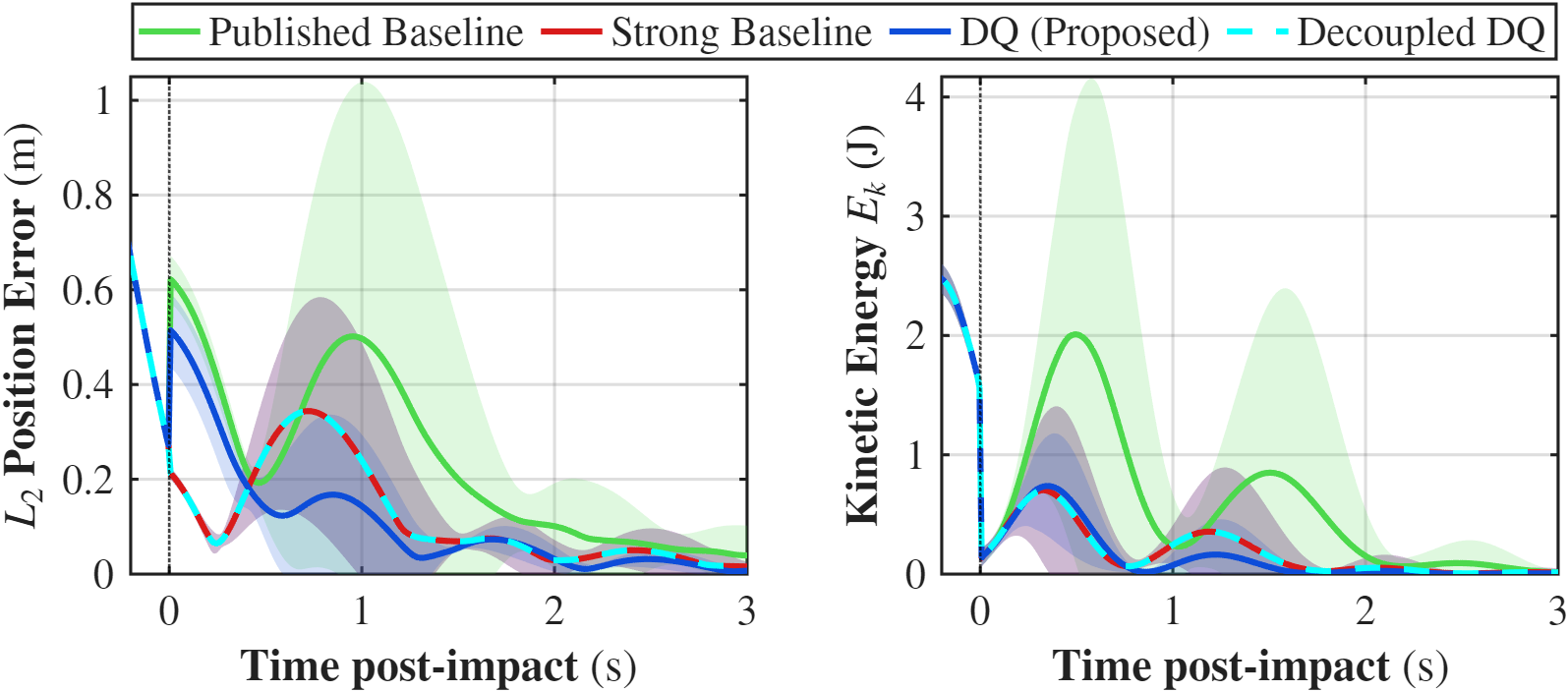}
        \caption{\textcolor{black}{Monte Carlo over initial $y-$position of the drone, $y\in[0,1]$}}
        \label{fig:mujoco_l2}
    \end{subfigure}
    \hfill
    \begin{subfigure}{1.0\linewidth}
        \centering
        \includegraphics[width=\linewidth]{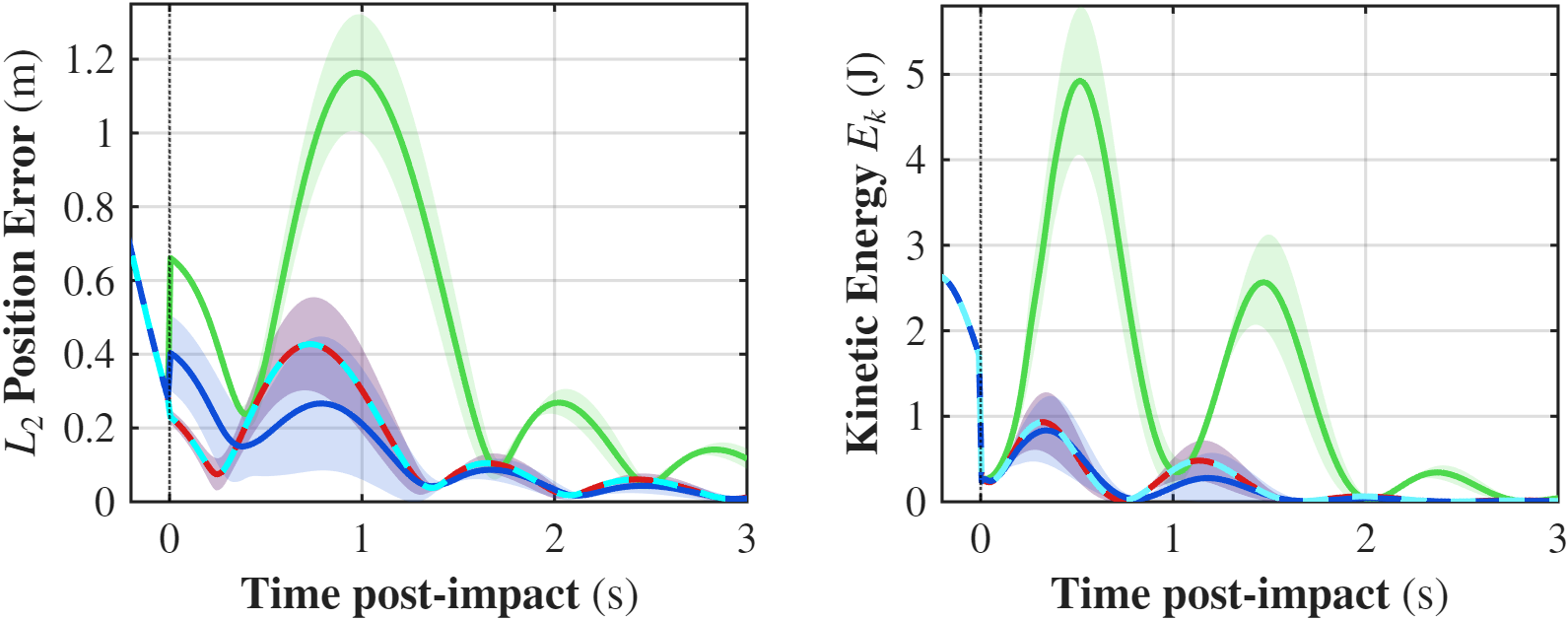}
        \caption{\textcolor{black}{Monte Carlo over different friction $\mu\in[0.1, 0.5]$.}}
        \label{fig:mujoco_ek}
    \end{subfigure}
    \caption{\textcolor{black}{Monte Carlo results ($N=50$) for the recovery phase.}}
    \label{fig:mujocosimenergy}
\end{figure}

We evaluate the performance of the DQ recovery framework of Sec. \ref{sec:recovery} \textcolor{black}{against two comparison points. (i) The published UAV collision-recovery baselines \cite{MarkMuller,Patnaik2021}, that apply translational admittance only with regulated post-impact attitude. (ii) A "strong" non-DQ baseline constructed as the extension of the drone literature, combining the classical friction decoupled matrix reset Eq. (\ref{eq:reset-SSM}) with a 6-DOF admittance shift in $\mathbb{R}^3 \times SO(3)$. To our knowledge, no published UAV collision-recovery method uses 6-DOF admittance shift. We also build the Decoupled-DQ by applying the assumption Eq.~(\ref{eq:assumption}) that, per Proposition 1, is equivalent to the Strong Baseline and is verified here with Fig.~\ref{fig:mujocosimenergy}}.\\
We perform computational analysis followed by an impulse contact model and  MuJoCo simulations for the recovery control, \textcolor{black}. All simulations are performed {with a UAV of 1 kg, thrust to weight ratio (TWR) of TWR=$3$, arm length of $r_c=0.25$ m and a wall at $(x_w,y_w)=(1,0)$ m.}

\vspace{0.05in}
\noindent \textbf{Computational Complexity Analysis.} We define the computational cost function $C(\cdot)$ as the total floating-point operations (FLOPs) needed to generate the reset map, assigning unit weights to additions and multiplications \cite{Wang2014}. We compare our approach against $SE(3)$ impulse determination strategies. The decoupled body-frame \textit{Matrix Formulation (MF)} requires explicit cross-products and matrix-vector multiplications, yielding $C(\text{MF})=69$. \textcolor{black}{A combined \textit{PQ} formulation requires extrinsic cross-products and quaternion vector rotations to resolve impact coupling, resulting in $C(\text{PQ}) \approx 65$.} Our \textit{DQ Formulation} bypasses these overheads, requiring $C(\text{DQ})=52$. This represents a 25\% FLOP reduction over the standard matrix baseline \textcolor{black}{and avoids the non-linear overhead of Lie exponential mappings.}\\
C++ benchmarks on a Raspberry Pi 5 validated these gains, demonstrating a 24\% execution time reduction. This latency drop expands the timing margin for concurrent autonomy threads on shared UAV hardware. Additionally, operating on 8-element DQ arrays rather than 12-element matrix-vector pairs maximizes cache coherency and reduces stack pressure, critical for Real-Time Operating System (RTOS) stability.

\vspace{0.05in}
\noindent \textbf{Impulse Contact Model Validation.}
The quadrotor is subjected to an idealized collision ($e=0.7$, $\mu=0.3$) using our coupled dual impulse model Eq.~(\ref{eq:DQM}) to evaluate the 3 controllers controller’s nominal transient performance (Published Baseline, Strong Baseline and Coupled DQ).
\begin{figure}[h]
    \centering
    \begin{subfigure}{0.49\linewidth}
        \centering
        \includegraphics[width=\linewidth]{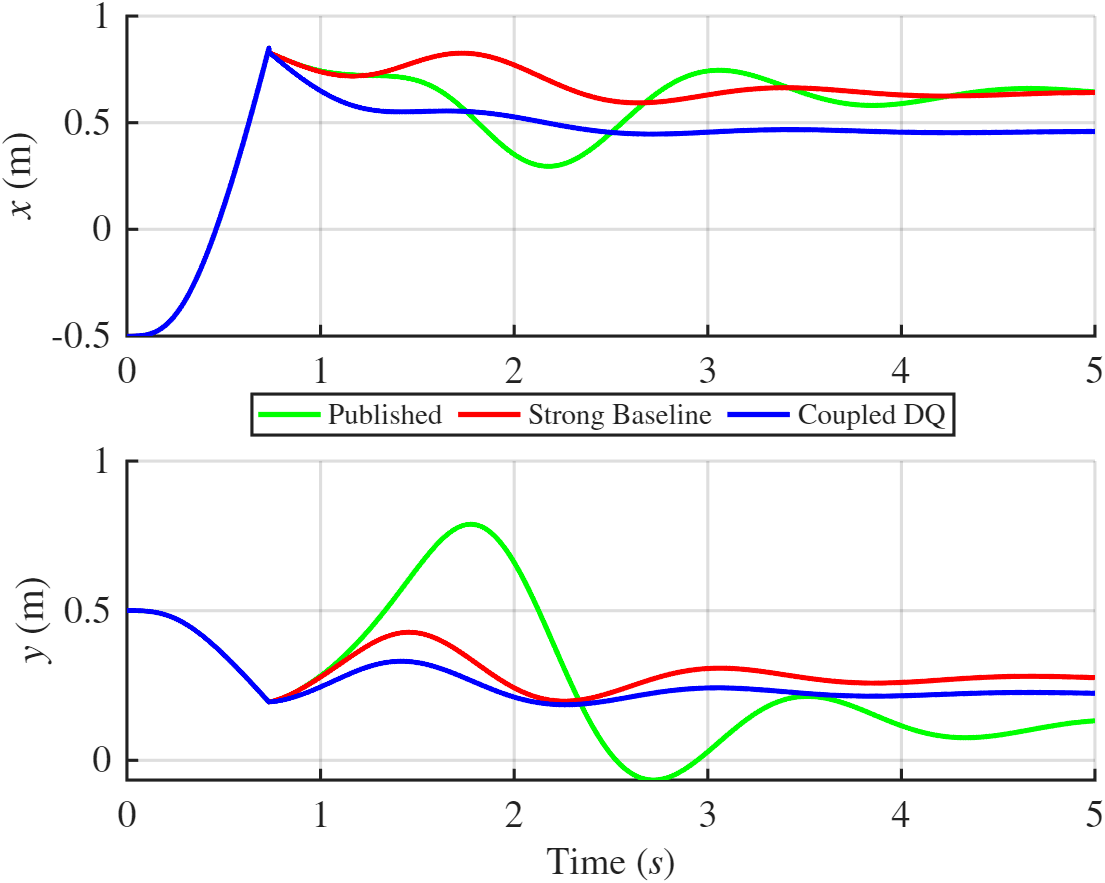}
        \caption{Position X/Y}
        \label{fig:Matlab_Position}
    \end{subfigure}
    \hfill
    \begin{subfigure}{0.49\linewidth}
        \centering
        \includegraphics[width=\linewidth]{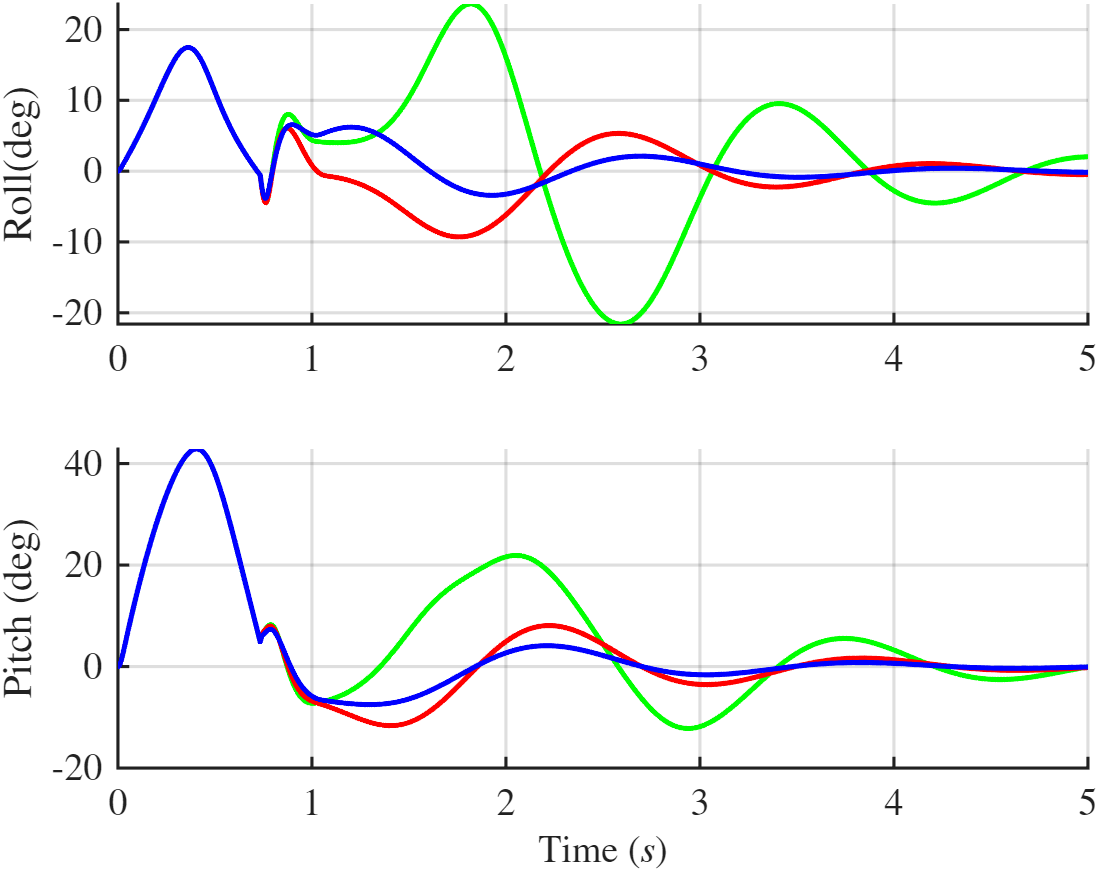}
        \caption{Orientation Pitch / Rill}
        \label{fig:Matlab_Orientation}
    \end{subfigure}
    \caption{Matlab simulations performed under impulse contact model.}
    \label{fig:matlabsim}
\end{figure}
As shown in Fig.~\ref{fig:matlabsim}, while the controllers achieve post-impact stability, their transient performance differs. The traditional Published baseline controller’s decoupled approach results in sustained attitude oscillations and longer settling time. The Strong Baseline and our DQ-method, by coupling linear and angular momentum, prioritizes total momentum dissipation. This geometric coupling enables more efficient energy dissipation and faster convergence to a stable 6-DOF recovery state. Additionally, we see the influence of the frictional coupling term in the DQ-Proposed method. By taking into account this frictional term, the resulting setpoint is more aligned with the system's true post-impact momentum.\\
In order to evaluate the controllers against a more realistic contact model, the next subsection use MuJoCo \cite{todorov2012mujoco} to further evaluate the controllers performances

\begin{figure}
    \vspace{0.5em}
    \centering
    \includegraphics[width=1.0\linewidth]{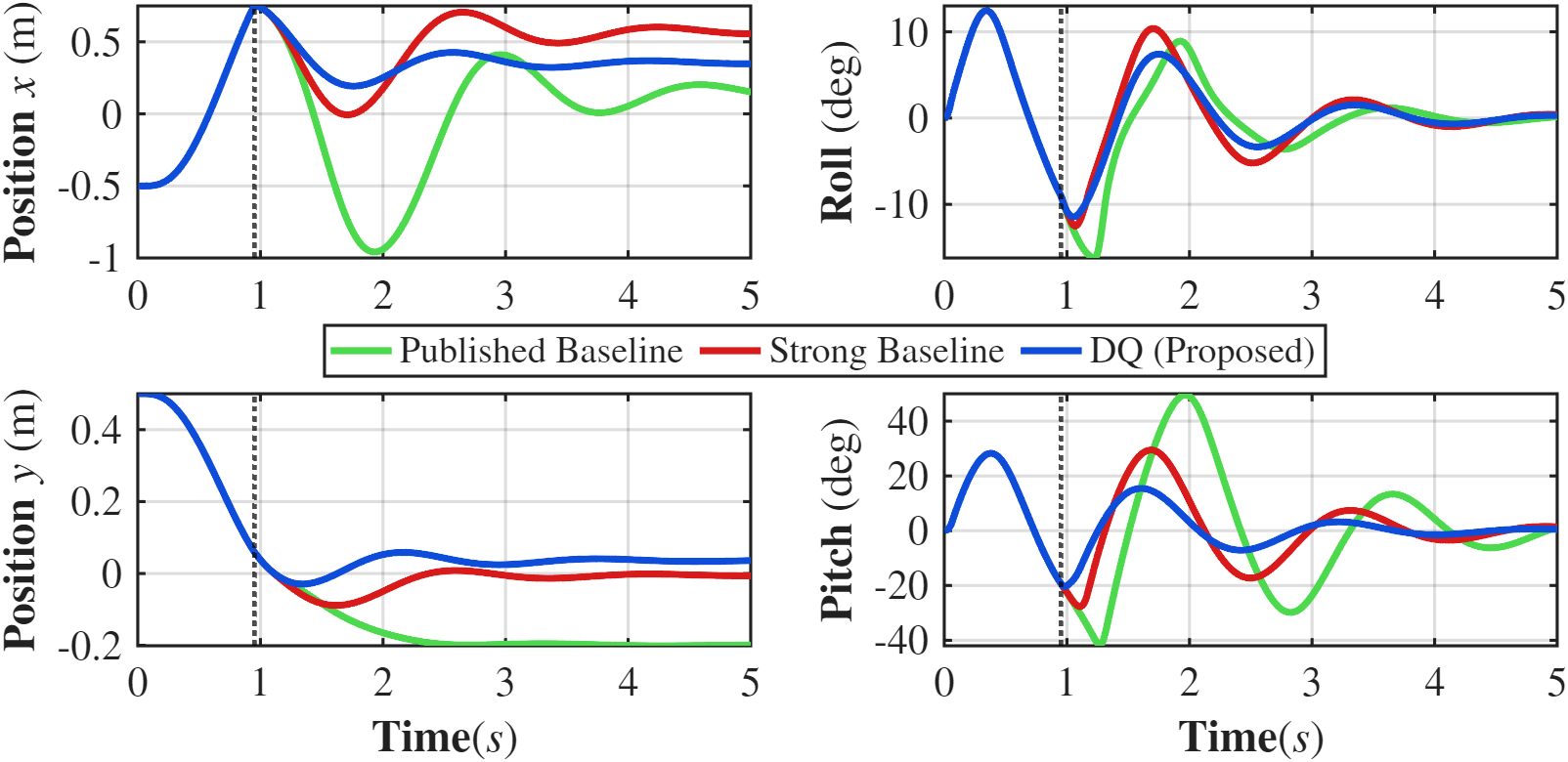}
    \caption{\textcolor{black}{MuJoCo results for an initial position $(x,y)=(-0.5,0.5)$ and friction $\mu=0.3$. Impact velocity $\approx2$ m/s.}}
    \label{fig:mujocosim}
\end{figure}

\vspace{0.05in}
\noindent \textbf{High-Fidelity Physics Simulation.} \textcolor{black}{In MuJoCo \cite{todorov2012mujoco}, soft-constraint LCP solvers break the decoupled assumption in \eqref{eq:assumption} by generating cross-coupled impact torques. This environment tests the recovery controller's ability against these unmodeled physical disturbances.} We compare the controllers using  Monte Carlo simulation measuring continuous $L_2$ position error \textcolor{black}{relative to the recovery setpoint to quantify spatial drift} and total kinetic energy $E_k$ \textcolor{black}{as it captures the coupled dissipation of both linear and angular velocities ($\frac{1}{2} m \| \textbf{v} \|^2 + \frac{1}{2}\boldsymbol{\omega}^T \mathbf{J} \boldsymbol{\omega}$)}, alongside a single-impact state response (Fig.~\ref{fig:mujocosim}). 
Figure~\ref{fig:mujocosim} shows \textcolor{black}{that the published baseline \cite{MarkMuller,Patnaik2021} loses lateral authority, suffering attitude oscillations. The difference in behavior between Strong Baseline and DQ confirms that the recovery quality is non-only driven by the 6-DOF admittance shift, but also by the ability of the proposed dual reset map to capture frictional coupling.}

\textcolor{black}{Monte Carlo simulations where performed over two different scenarios. One over impact angles, by varying the initial $y-$position of the drone ($y\in[0,1]$)
(Fig.~\ref{fig:mujoco_l2}). The other performed over the friction coefficients (Fig.~\ref{fig:mujoco_ek}), by varying $\mu\in[0.1, 0.5]$; $\mu \leq 0.5$ chosen to avoid the sticking regime of Coulomb friction law that the impulse formulation doesn't tackle. First, the DQ method improves the $L_2$ RMSE by 50.8--75.1\% and the $E_k$ RMSE by
68.7--85.0\% over the published baselines
\cite{MarkMuller,Patnaik2021} (Table~\ref{tab:metrics}), maintaining bounded lateral deviation and smooth kinetic-energy decay. This gap reflects the inclusion of
angular admittance, absent from existing UAV recovery methods. Second,
the DQ-decoupled case, in which the cross-coupling term of
Eq.~(\ref{eq:DQM}) is neglected (Eq.~(\ref{eq:assumption})), coincides with the strong baseline,
providing an empirical verification of Proposition~1 and isolating the
cross-coupling correction as the source of the residual difference
between the DQ controller and the strong baseline. Relative to the
strong baseline, the coupled DQ recovery reduces the $L_2$ RMSE by
12.4\% and the $E_k$ RMSE by 4.0\% in the angle sweep, and by 24.7\%
and 12.7\% respectively in the friction sweep
(Table~\ref{tab:metrics}). The improvement is larger under higher
friction, where the cross-coupling term is most active.}

\begin{table}
\centering
\caption{Improvement of the proposed DQ controller using post-impact recovery metrics from Monte Carlo simulation ($N=50$) against two baseline methods.}
\label{tab:metrics}
\begin{tabular}{llcc}
\toprule
\textbf{Variation} & \textbf{Metric} & \textbf{vs Published} & \textbf{vs Strong Baseline} \\
\midrule
\multirow{2}{*}{\rotatebox[origin=c]{0}{\textbf{Angle}}}
 & $L_2$ RMSE (m)     & 50.8\% & 12.4\% \\
 & $E_k$ RMSE (J)     & 68.7\% & 4.0\%  \\
\midrule
\multirow{2}{*}{\rotatebox[origin=c]{0}{\textbf{Friction}}}
 & $L_2$ RMSE (m)     & 75.1\% & 24.7\% \\
 & $E_k$ RMSE (J)     & 85.0\% & 12.7\% \\
\bottomrule
\end{tabular}
\end{table}

%% file: 05-sectionV.tex
\section{Conclusion}
This letter presents a dual quaternion reset map for rigid-body aerial impacts on the $SE(3)$ manifold. \textcolor{black}{The closed-form impulse update retains the cross-coupling between normal and tangential components in a close expression and recovers the classical decoupled formulation under Proposition 1. Monte Carlo simulations show that the resulting controller outperforms published admittance baselines and decoupled friction formulations, while achieving a reduced computational load. The proposed reset map is applicable to a wide range of contact-rich applications such as aerial manipulation as well as rendezvous, proximity operations and docking (RPOD) in space, where the cross-coupling is critical for mission success. A shared limitation with classical impulse models is our formulation's assumption of sliding contact conditions. Future work will integrate sensing for contact wrench estimation and embed the reset map into impact-aware drone path planning.}

%% file: 06-appendix.tex
\appendix
\subsection{Extended Hybrid Lyapunov Stability Proof}
\label{sec:Lyapunov}

This appendix provides the complete mathematical derivation of the
hybrid Lyapunov stability conditions for the proposed dual quaternion
closed-loop recovery system.

\vspace{0.05in}
\noindent \textbf{System Definition and Lyapunov Candidate}: We define the hybrid system state as $x = (\hat{\mathbf{q}}_e,
\hat{\boldsymbol{\xi}})$, where $\hat{\mathbf{q}}_e \in \mathcal{H}_d$
is the dual quaternion pose error and $\hat{\boldsymbol{\xi}} \in
\mathcal{H}_d^p$ is the body twist. The hybrid dynamical system is
defined as $\mathcal{H} = (F, \mathcal{C}, \mathcal{D}, \hat{\mathcal{R}})$
following the standard framework of~\cite{goebel2012hybrid}.

We consider the candidate Lyapunov function
\begin{equation}
V(x) = \underbrace{2k_q(1 - q_{e}^w) +
\frac{1}{2}k_p \|\mathbf{p}_e\|^2}_{V_{pos}}
+ \underbrace{\frac{1}{2}\langle \hat{\boldsymbol{\xi}},
\mathcal{M}(\hat{\boldsymbol{\xi}}) \rangle}_{V_{kin}},
\end{equation}
where $q_e^w \in \mathbb{R}$ is the scalar real part of the unit
quaternion $\mathbf{q}_e \in \mathbb{S}^3$, $\mathbf{p}_e \in
\mathbb{R}^3$ is the translational error, and $\mathcal{M} \succ 0$ is
the dual inertia operator defined in Eq.~(6) of the main letter.
Both $V_{pos}$ and $V_{kin}$ are non-negative, so $V(x) \geq 0$ $\forall x$.\\
Positive definiteness of $V_{pos}$ with respect to the target set
$\mathcal{A} = \{x \mid \hat{\mathbf{q}}_e = \hat{\mathbf{1}},\,
\hat{\boldsymbol{\xi}} = \hat{\mathbf{0}}\}$ requires that the
shortest-path quaternion convention be enforced. Specifically, the
scalar part $q_e^w$ is constrained to be non-negative prior to each
control evaluation by mapping $\hat{\mathbf{q}}_e \leftarrow
-\hat{\mathbf{q}}_e$ whenever $q_e^w < 0$.\\
Under this convention, $2k_q(1 - q_e^w) = 0 \Leftrightarrow q_e^w = 1$, i.e.,
$\hat{\mathbf{q}}_e = \hat{\mathbf{1}}$, and $V_{pos}$ is positive
definite on $\mathcal{A}$ as required. The kinetic term satisfies
$\frac{1}{2}\langle \hat{\boldsymbol{\xi}}, \mathcal{M}(
\hat{\boldsymbol{\xi}})\rangle = 0 \Leftrightarrow \hat{\boldsymbol{\xi}} = \hat{\mathbf{0}}$, since $\mathcal{M}
\succ 0$. Consequently, $V(x) = 0 \Leftrightarrow x \in \mathcal{A}$,
confirming that $V$ is a valid Lyapunov candidate.

\vspace{0.05in}
\noindent 
\textbf{Continuous Dynamics (Flow Set $\mathcal{C}$)}: During free flight the state lies in the flow set $\mathcal{C} =
\{x \mid \phi(\hat{\mathbf{q}}) > 0\}$, where $\phi$ denotes the
signed distance to the contact surface. The time derivative of $V$
decomposes as $\dot{V} = \dot{V}_{pos} + \dot{V}_{kin}$.

Taking the time derivative of $V_{pos}$ and applying the dual
quaternion kinematic identity $\dot{\hat{\mathbf{q}}}_e =
\frac{1}{2}\hat{\mathbf{q}}_e \otimes \hat{\boldsymbol{\xi}}$
yields the inner product with the generalized pose-error gradient
$\hat{\mathbf{e}}$,
\begin{equation}
\dot{V}_{pos} = \langle \hat{\mathbf{e}},\, \hat{\boldsymbol{\xi}}
\rangle, \qquad
\hat{\mathbf{e}} \triangleq k_q \mathbf{q}_{e}^v
+ \varepsilon k_p \mathbf{p}_e,
\end{equation}
where $\mathbf{q}_e^v \in \mathbb{R}^3$ is the vector part of
$\mathbf{q}_e$.\\
Differentiating the kinetic term and substituting the dual dynamics
of Eq.~(4) gives
\begin{equation}
\dot{V}_{kin}
= \langle \hat{\boldsymbol{\xi}},\,
\mathcal{M}(\dot{\hat{\boldsymbol{\xi}}}) \rangle
= \langle \hat{\boldsymbol{\xi}},\,
\hat{\mathbf{F}}_a + \hat{\mathbf{F}}_g
- \hat{\boldsymbol{\xi}} \times^{*} \mathcal{M}(\hat{\boldsymbol{\xi}})
\rangle.
\end{equation}
By the skew-symmetry of the dual co-adjoint operator, the gyroscopic
term satisfies $\langle \hat{\boldsymbol{\xi}},\,
\hat{\boldsymbol{\xi}} \times^{*} \mathcal{M}(\hat{\boldsymbol{\xi}})
\rangle = 0$ identically. Substituting the control law of Eq.~(23),
\begin{equation}
\hat{\mathbf{F}}_a = -\hat{\mathbf{F}}_g
+ \hat{\boldsymbol{\xi}} \times^{*} \mathcal{M}(\hat{\boldsymbol{\xi}})
- \hat{\mathbf{e}}
- \mathbf{K}_d \circ \hat{\boldsymbol{\xi}},
\end{equation}
into $\dot{V}_{kin}$, and summing with $\dot{V}_{pos}$, produces
\begin{equation}
\dot{V}
= \langle \hat{\mathbf{e}},\, \hat{\boldsymbol{\xi}} \rangle
+ \langle \hat{\boldsymbol{\xi}},\,
{-\hat{\mathbf{e}} - \mathbf{K}_d \circ \hat{\boldsymbol{\xi}}} \rangle.
\end{equation}
The potential gradient terms $\langle \hat{\mathbf{e}},
\hat{\boldsymbol{\xi}} \rangle$ cancel exactly, leaving
\begin{equation}
\dot{V} = -\langle \hat{\boldsymbol{\xi}},\,
\mathbf{K}_d \circ \hat{\boldsymbol{\xi}} \rangle \leq 0.
\end{equation}
Because $\mathbf{K}_d \succ 0$, this inequality holds for all $x \in
\mathcal{C}$, establishing that energy is non-increasing along
continuous flows.

\vspace{0.05in}
\noindent 
\textbf{Impact Dynamics (Jump Set $\mathcal{D}$)}: At the instant of impact the state enters the jump set $\mathcal{D}
= \{x \mid \phi(\hat{\mathbf{q}}) \leq 0\}$. The total change in the
Lyapunov function across the discrete event is $\Delta V =
V(x^+) - V(x^-) = \Delta V_{pos} + \Delta V_{kin}$. The two
contributions are analyzed separately.\\
\vspace{0.05in}
\noindent \textit{Kinetic energy dissipation.} The instantaneous change in kinetic energy is
\begin{equation}
\Delta V_{kin}
= \frac{1}{2}\langle \hat{\boldsymbol{\xi}}^+,\,
\mathcal{M}(\hat{\boldsymbol{\xi}}^+) \rangle
- \frac{1}{2}\langle \hat{\boldsymbol{\xi}}^-,\,
\mathcal{M}(\hat{\boldsymbol{\xi}}^-) \rangle.
\end{equation}
Using the dual momentum update $\mathcal{M}(\hat{\boldsymbol{\xi}}^+)
= \mathcal{M}(\hat{\boldsymbol{\xi}}^-) + \hat{\mathcal{W}}$ from
Eq.~(8) and expanding, this becomes
\begin{equation}
\Delta V_{kin}
= \langle \hat{\boldsymbol{\xi}}^-,\, \hat{\mathcal{W}} \rangle
+ \frac{1}{2}\langle \hat{\mathcal{W}},\,
\mathcal{M}^{-1}(\hat{\mathcal{W}}) \rangle.
\end{equation}
Substituting the normal wrench component $\hat{\mathcal{W}}_n =
\Lambda \hat{\mathbf{s}}_n$ from Eq.~(11) and applying Newton's
restitution constraint $\langle \hat{\mathbf{s}}_n,
\hat{\boldsymbol{\xi}}^+ \rangle = -e\langle \hat{\mathbf{s}}_n,
\hat{\boldsymbol{\xi}}^- \rangle$ to eliminate $\langle
\hat{\boldsymbol{\xi}}^-, \hat{\mathbf{s}}_n \rangle$ in terms of
$\Lambda$, the normal contribution to the kinetic energy change is
\begin{equation}
\Delta V_{kin,n}
= -\frac{1}{2}\Lambda^2
\langle \hat{\mathbf{s}}_n,\, \mathcal{M}^{-1}(\hat{\mathbf{s}}_n)
\rangle \left(\frac{1-e}{1+e}\right).
\end{equation}
Since $\mathcal{M} \succ 0$ implies $\langle \hat{\mathbf{s}}_n,
\mathcal{M}^{-1}(\hat{\mathbf{s}}_n)\rangle > 0$ and $e \in [0,1)$
ensures $\frac{1-e}{1+e} > 0$, this term is strictly negative for
any non-trivial impact ($\Lambda \neq 0$). The Coulomb friction model
with coefficient $\mu \geq 0$ additionally ensures $\Delta V_{kin,t}
\leq 0$ for the tangential component. The total kinetic energy change
is therefore strictly dissipative,
\begin{equation}
\Delta V_{kin} = -E_{diss} < 0,
\end{equation}
where $E_{diss} > 0$ collects both normal and tangential contributions.\\

\vspace{0.05in}
\noindent \textit{Potential energy change.} Under Assumption~1, the UAV configuration is continuous across the impact event, so $\hat{\mathbf{q}}^+ = \hat{\mathbf{q}}^-$.
The reference setpoint is updated according to Eq.~(22a),
\begin{equation}
\hat{\mathbf{q}}_d^+ = \hat{\mathbf{q}}_\Delta \otimes
\hat{\mathbf{q}}_d^-,
\end{equation}
where $\hat{\mathbf{q}}_\Delta = \exp(\hat{\boldsymbol{\delta}}/2)$
is the bounded pose shift induced by the admittance map. The
post-impact pose error is
\begin{equation}
\hat{\mathbf{q}}_e^+ = \hat{\mathbf{q}}_d^{+\,*} \otimes
\hat{\mathbf{q}}
= \hat{\mathbf{q}}_d^{-\,*} \otimes \hat{\mathbf{q}}_\Delta^{*}
\otimes \hat{\mathbf{q}}.
\end{equation}
Because $V_{pos}$ depends on $\hat{\mathbf{q}}_e$ through the
non-negative quantities $1 - q_e^w$ and $\|\mathbf{p}_e\|^2$, and
since composing with $\hat{\mathbf{q}}_\Delta^*$ on the right
introduces an additional displacement bounded by
$V_{pos}(\hat{\mathbf{q}}_\Delta)$, it follows from the
sub-multiplicativity of the dual quaternion norm that
\begin{equation}
\Delta V_{pos} = V_{pos}(\hat{\mathbf{q}}_e^+)
- V_{pos}(\hat{\mathbf{q}}_e^-)
\leq V_{pos}(\hat{\mathbf{q}}_\Delta).
\end{equation}
The total Lyapunov change across the jump is therefore bounded by
\begin{equation}
\Delta V \leq V_{pos}(\hat{\mathbf{q}}_\Delta) - E_{diss}.
\end{equation}
For $\Delta V < 0$ it is sufficient to require
$V_{pos}(\hat{\mathbf{q}}_\Delta) < E_{diss}$, which is guaranteed
by the gain condition derived in the following subsection.\\

\vspace{0.05in}
\noindent \textit{Bounding $V_{pos}(\hat{\mathbf{q}}_\Delta)$.} To obtain an explicit bound on $V_{pos}(\hat{\mathbf{q}}_\Delta)$,
we bound the scalar real part of $\hat{\mathbf{q}}_\Delta =
\exp(\hat{\boldsymbol{\delta}}/2)$ using the exact trigonometric
inequality $1 - \cos\theta \leq \frac{\theta^2}{2}$, which holds
for all $\theta \in \mathbb{R}$. With $\theta =
\|\boldsymbol{\delta}_\omega\|/2$, this gives
\begin{equation}
1 - q_\Delta^w = 1 - \cos\!\left(\frac{\|\boldsymbol{\delta}_\omega\|}{2}
\right) \leq \frac{\|\boldsymbol{\delta}_\omega\|^2}{8}.
\end{equation}
For the translational component, the dual exponential map yields
$\mathbf{p}_\Delta = \boldsymbol{\delta}_v$ at leading order, with
higher-order terms that are non-negative, so
$\frac{1}{2}k_p\|\mathbf{p}_\Delta\|^2 \leq
\frac{1}{2}k_p\|\boldsymbol{\delta}_v\|^2$ provides a valid upper
bound. Substituting $\hat{\boldsymbol{\delta}} = \boldsymbol{\Gamma}_\omega
\boldsymbol{\omega}^+ + \varepsilon \boldsymbol{\Gamma}_v
\mathbf{v}_B^+$ from Eq.~(21),
\begin{align}
V_{pos}(\hat{\mathbf{q}}_\Delta)
&\leq 2k_q \cdot \frac{\|\boldsymbol{\Gamma}_\omega
\boldsymbol{\omega}^+\|^2}{8}
+ \frac{1}{2}k_p
\|\boldsymbol{\Gamma}_v \mathbf{v}_B^+\|^2 \notag \\
&= \frac{1}{4}k_q\|\boldsymbol{\Gamma}_\omega \boldsymbol{\omega}^+\|^2
+ \frac{1}{2}k_p\|\boldsymbol{\Gamma}_v \mathbf{v}_B^+\|^2.
\end{align}
The sufficient condition $V_{pos}(\hat{\mathbf{q}}_\Delta) < E_{diss}$
therefore becomes
\begin{equation}
\frac{1}{4}k_q\|\boldsymbol{\Gamma}_\omega \boldsymbol{\omega}^+\|^2
+ \frac{1}{2}k_p\|\boldsymbol{\Gamma}_v \mathbf{v}_B^+\|^2
< E_{diss}.
\end{equation}
Adopting the more conservative global bound
$E(\hat{\mathbf{q}}_e^-) = E_{diss} + V_{pos}^- \geq E_{diss}$
yields the condition stated in Eq.~(29) of the main letter.\\

\vspace{0.05in}
\noindent \textit{Admittance Gain Synthesis.} To derive explicit design bounds, we introduce a budget-allocation
parameter $\alpha \in (0,1)$ that partitions the energy bound between
the rotational and translational degrees of freedom,
\begin{subequations}
\begin{align}
\frac{1}{4}k_q\|\boldsymbol{\Gamma}_\omega \boldsymbol{\omega}^+\|^2
&< \alpha\, E(\hat{\mathbf{q}}_e^-), \\
\frac{1}{2}k_p\|\boldsymbol{\Gamma}_v \mathbf{v}_B^+\|^2
&< (1-\alpha)\, E(\hat{\mathbf{q}}_e^-).
\end{align}
\end{subequations}
Isolating the spectral norms of the gain matrices yields the
explicit upper bounds
\begin{equation}
\|\boldsymbol{\Gamma}_\omega\|_2
< \sqrt{\frac{4\alpha\, E(\hat{\mathbf{q}}_e^-)}
{k_q\|\boldsymbol{\omega}^+\|^2}},
\qquad
\|\boldsymbol{\Gamma}_v\|_2
< \sqrt{\frac{2(1-\alpha)\, E(\hat{\mathbf{q}}_e^-)}
{k_p\|\mathbf{v}_B^+\|^2}}.
\end{equation}
The bounds depend on the pre-impact energy $E(\hat{\mathbf{q}}_e^-)$,
which is a runtime quantity determined by the pre-impact twist and
pose error. In practice, the gains $(\boldsymbol{\Gamma}_\omega,
\boldsymbol{\Gamma}_v)$ are selected offline by evaluating the bounds
at the worst-case anticipated pre-impact velocity. Any collision whose
pre-impact conditions fall within that envelope then satisfies
$\Delta V < 0$ by construction.

Since $\dot{V} \leq 0$ for all $x \in \mathcal{C}$ and $\Delta V < 0$
for all $x \in \mathcal{D}$ under the gain bounds above, the target
set $\mathcal{A}$ is Lyapunov stable for the hybrid closed-loop
system $\mathcal{H}$ in the sense of~\cite{goebel2012hybrid}.
\hfill $\blacksquare$